\documentclass[conference]{IEEEtran}
\IEEEoverridecommandlockouts

\usepackage{cite}
\usepackage{amsmath,amssymb,amsfonts}
\usepackage{algorithmic}
\usepackage{graphicx}
\usepackage{textcomp}
\usepackage{xcolor}
\usepackage{wrapfig}
\usepackage{multirow}
\usepackage{tabularx,booktabs}
\usepackage{tabularray}
\usepackage{arydshln}
\usepackage{hhline}
\usepackage{nicematrix}

\usepackage{caption}
\usepackage{subcaption}
\usepackage{bbm}
\usepackage{hyperref}

\usepackage[a4paper, total={184mm,239mm}]{geometry}
\def\BibTeX{{\rm B\kern-.05em{\sc i\kern-.025em b}\kern-.08em
    T\kern-.1667em\lower.7ex\hbox{E}\kern-.125emX}}
\begin{document}

\title{MD-SNN: Membrane Potential-aware Distillation on Quantized Spiking Neural Network
}

\author{Donghyun Lee$^1$, Abhishek Moitra$^2$, Youngeun Kim$^3\dagger$, Ruokai Yin$^4$, Priyadarshini Panda$^1$ \\
$^1$University of Southern California, $^2$Washington State University, $^3$Amazon, $^4$Yale University\\
donghyun.lee.1@usc.edu}


\maketitle

\begingroup
\renewcommand\thefootnote{$\dagger$}
\footnotetext{Work done prior to joining Amazon.}
\endgroup

\begin{abstract}
Spiking Neural Networks (SNNs) offer a promising and energy-efficient alternative to conventional neural networks, thanks to their sparse binary activation. However, they face challenges regarding memory and computation overhead due to complex spatio-temporal dynamics and the necessity for multiple backpropagation computations across timesteps during training. To mitigate this overhead, compression techniques such as quantization are applied to SNNs. Yet, naively applying quantization to SNNs introduces a mismatch in membrane potential, a crucial factor for the firing of spikes, resulting in accuracy degradation. In this paper, we introduce Membrane-aware Distillation on quantized Spiking Neural Network (MD-SNN), which leverages membrane potential to mitigate discrepancies after weight, membrane potential, and batch normalization quantization. To our knowledge, this study represents the first application of membrane potential knowledge distillation in SNNs. We validate our approach on various datasets, including CIFAR10, CIFAR100, N-Caltech101, and TinyImageNet, demonstrating its effectiveness for both static and dynamic data scenarios. Furthermore, for hardware efficiency, we evaluate the MD-SNN with SpikeSim platform, finding that MD-SNNs achieve 14.85$\times$ lower energy-delay-area product (EDAP), 2.64$\times$ higher TOPS/W, and 6.19$\times$ higher TOPS/mm$^2$ compared to floating point SNNs at iso-accuracy on N-Caltech101 dataset. Code is available at \href{https://github.com/Intelligent-Computing-Lab-Panda/MD-SNN}{Github}.
\end{abstract}


\vspace{-3mm}
\section{Introduction}

In recent years, Spiking Neural Networks (SNNs) have emerged as a compelling energy-efficient alternative to conventional Artificial Neural Networks (ANNs) \cite{maass1997networks,roy2019towards}. Distinguished by their event-driven computation and sparse binary spike communication across temporal dimension, SNNs offer a biologically-inspired paradigm for neural information processing. This unique computational model not only promises significant energy savings but also enables deployment across diverse hardware platforms, from neuromorphic chips to edge devices \cite{akopyan2015truenorth,davies2018loihi,yin2022sata, yin2024loas,li2025deterministic}.

The pursuit of energy-efficient SNNs has spawned two primary training methodologies: ANN-to-SNN conversion \cite{diehl2015fast,han2020deep,li2021free,deng2021optimal, bu2023optimal} and direct training via backpropagation with surrogate gradients \cite{wu2018spatio,shrestha2018slayer,zheng2021going,lee2025spiking}. While backpropagation-based approaches have achieved state-of-the-art accuracy with fewer timesteps, they impose substantial memory overhead due to the storage requirement for temporal activations during training \cite{yin2022sata}. To address these computational challenges, researchers have developed various compression techniques, including quantization \cite{putra2021q,li2022quantization,yinmint}, pruning \cite{yin2023workload}, Knowledge Distillation (KD)~\cite{kushawaha2021distilling,xu2023constructing,guo2023joint}, and tensor decomposition \cite{lee2024tt}.

\begin{figure}
    \centering
    \includegraphics[width=90mm]{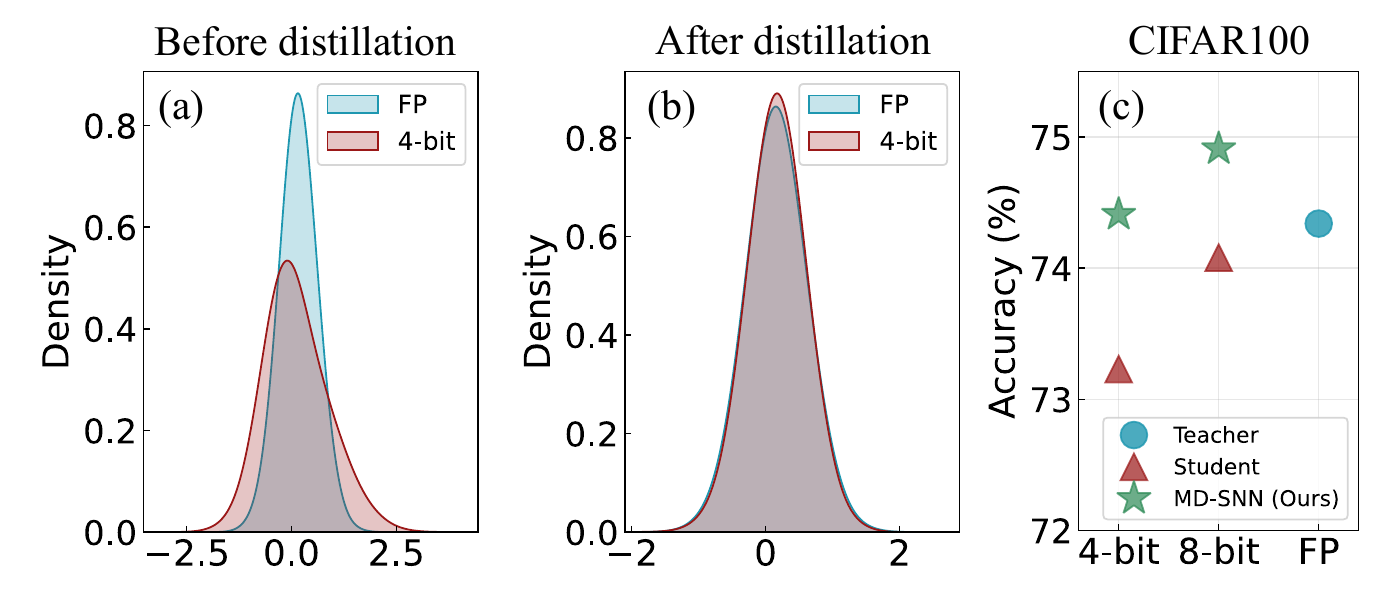}
    \vspace{-6.5mm}
    \caption{Impact of knowledge distillation on membrane potential distributions. (a) Before distillation: significant mismatch between Floating-Point (FP) and 4-bit quantized models. (b) After distillation: aligned distributions through MD-SNN. (c) Performance comparison on CIFAR100, where MD-SNN (green star) surpasses both the quantized student and FP teacher models.}
    \label{fig:mem_mismatch}
    \vspace{-6mm}
\end{figure}

Among these techniques, quantization has proven particularly effective in reducing both memory footprint and computational complexity. By reducing the bit-precision of weights and membrane potentials, quantization methods have successfully compressed SNNs while maintaining competitive accuracy \cite{putra2021q,li2022quantization}. For instance, recent work has demonstrated that SNNs can operate with weights and membrane potentials quantized to as low as 2-bit integer precision with minimal accuracy loss \cite{yinmint}. However, aggressive quantization introduces a critical challenge: \textit{significant distortion in membrane potential distributions across timesteps, which directly impacts spike generation and network performance}.


Fig.~\ref{fig:mem_mismatch}(a) illustrates the critical challenge: membrane potential distributions between Floating-Point (FP) and 4-bit quantized models exhibit substantial divergence, disrupting the precise spike timing essential for SNN information encoding. This divergence is particularly problematic because membrane potentials, the continuous state variables that integrate temporal inputs and govern spike generation, encode rich dynamical information absent from binary spike outputs. Our KD approach successfully aligns these distributions (Fig.~\ref{fig:mem_mismatch}(b)), achieving performance that surpasses even the full-precision teacher (Fig.~\ref{fig:mem_mismatch}(c)). While KD has been explored for SNNs, existing methods exhibit fundamental limitations. Approaches using ANN teachers \cite{xu2023constructing,guo2023joint} risk losing spike-specific temporal dynamics, while SNN-based methods focusing solely on output spikes \cite{kushawaha2021distilling} overlook the informative membrane potential trajectories. By directly transferring membrane distributions, our work preserves the complete spatiotemporal dynamics critical for accurate spike generation in quantized networks.

In this paper, we introduce Membrane-aware Distillation for quantized Spiking Neural Networks (MD-SNN), a novel knowledge distillation framework that explicitly leverages membrane potential dynamics to guide quantized student networks. As illustrated in Fig.~\ref{fig:main}(a), MD-SNN employs dual distillation pathways, i.e., logit and membrane potential, between a full-precision SNN teacher and a quantized student. The key innovation lies in recognizing membrane potentials as feature representations analogous to hidden activations in ANNs, but with additional temporal dynamics unique to SNNs. Our approach makes several critical design choices. First, we employ Kullback-Leibler (KL) divergence rather than the conventional Mean Squared Error (MSE) for membrane distillation. Our empirical analysis reveals that MSE leads to excessive spike generation, creating redundant information flow that degrades performance. KL divergence, by treating membrane distributions probabilistically, better preserves the relative importance of different membrane states. Second, we introduce a versatile teacher framework where a single teacher trained with $T$ timesteps can effectively guide multiple students operating at timestep $t \leq T$, diminishing training costs across different temporal configurations. The main contributions of this work are as:

\begin{itemize}
    \item We present the first knowledge distillation method that explicitly transfers membrane potential dynamics between SNN models, from full-precision teachers to quantized students. Our temporal and spatial (group-wise) distillation design achieves an optimal balance between effectiveness and computational efficiency.
    
    \item Our framework enables a single teacher model with timestep $T$ to guide students with varying timesteps ($t \leq T$), reducing the training overhead by 30\% for multi-configuration deployments.
    
    
    \item Extensive experiments on static (CIFAR-10/100, TinyImageNet) and neuromorphic (N-Caltech101) datasets demonstrate that MD-SNN consistently outperforms existing quantization methods, achieving hardware efficiency with up to 14.85$\times$ EDAP reduction while maintaining competitive accuracy.
\end{itemize}

\vspace{-3mm}
\section{Related Work}
\vspace{-1mm}
\subsection{Spiking Neural Network}
\vspace{-1mm}
The advancement of SNNs has spurred an extensive investigation into methods for effectively training and deploying SNNs. Particularly, the training method of SNNs can be divided into two categories, i.e., ANN-to-SNN conversion and backpropagation with surrogate gradients. Within the domain of SNN research, one prominent avenue is the conversion of conventional ANNs into their spiking counterparts. This approach involves the transformation of pre-trained ANNs, typically designed for continuous value operations, into SNN configurations. The primary objective is to capitalize on the wealth of knowledge cultivated in the ANN domain to expedite and enhance the training of SNNs. Eminent researchers in this field, including Diehl et al. \cite{diehl2015fast,han2020deep,li2021free,deng2021optimal}, have made substantial contributions by elucidating the conversion process and demonstrating the potential of ANNs as instructive mentors for SNNs.

Backpropagation, a well-established method for training conventional ANNs, has found application in the context of SNNs, albeit with requisite modifications to accommodate the discrete nature of spike events. An influential facet of this research realm is the utilization of surrogate gradients, which streamline the integration of backpropagation techniques for SNN training. \cite{wu2018spatio, shrestha2018slayer,zheng2021going, lee2025spiking} have significantly contributed to this domain by offering valuable insights into the training of SNNs.

\vspace{-0.5mm}
\subsection{Efficient methods for compressing SNNs}
The training of SNNs involves computationally intensive backward computations, particularly in surrogate gradient calculations. To mitigate this computational burden, a range of efficient methods have been introduced, all focused on reducing memory and computation costs such as KD \cite{xu2023constructing, kushawaha2021distilling, guo2023joint}, quantization \cite{putra2021q, li2022quantization, yinmint}, pruning \cite{yin2023workload}, and tensor decomposition \cite{lee2024tt}. In terms of KD, Xu et al.~\cite{xu2023constructing} propose an ANN-based teacher model that imparts its knowledge to an SNN-based student model, resulting in accelerated convergence. Kushawaha et al.~\cite{kushawaha2021distilling} take a different route, employing an SNN-based teacher model, which shares its spike distribution knowledge with the student model to stabilize the training process. Guo et al. \cite{guo2023joint} introduce a novel KD approach by transferring knowledge from an ANN teacher model to an SNN student model using both logit and feature distillation. Additionally, they enforce constraints to ensure that the ANN and SNN models share the same singular vectors during the training process, resulting in improved accuracy. These KD strategies have demonstrated their effectiveness in enabling small SNN models to achieve performance levels comparable to those of the teacher model. Other approaches, such as pruning and tensor decomposition, have been introduced to make SNN architectures more hardware-friendly. Unstructured pruning, guided by the Lottery Ticket Hypothesis (LTH) \cite{kim2022exploring}, has been effectively employed in SNNs \cite{yin2023workload}. Moreover, tensor decomposition techniques, particularly the Tensor Train (TT) decomposition, have been harnessed to optimize SNNs, leading to reductions in weight count, training energy expenditure, and training time \cite{lee2024tt}.

\vspace{-3mm}
\section{Preliminary}
\subsection{Leaky Integrate-and-Fire Neuron}
As a fundamental building block of SNNs, the Leaky Integrate-and-Fire (LIF) neuron has emerged as an important component for energy-efficient computation. The LIF neuron is a non-linear activation function that determines whether neurons fire spikes as follows:
\begin{align}
    \mathbf{u}^{t+1}_l = \tau_m \mathbf{u}^{t}_l&+\mathbf{W}_lf(\mathbf{u}^{t}_{l-1}) \label{lif1} \\
    f(\mathbf{u}_l^t) =& \mathbbm{1}_{\mathbf{u}_l^t > V_{th}} \label{lif2}
\end{align}
\noindent where, $\mathbf{u}^t_l$ is the membrane potential in $l$-th layer at timestep $t$, $\tau_m\in(0,1]$ is the leaky factor for membrane potential leakage, $\mathbf{W}$ is the weight, and $f(\cdot)$ is the LIF function with firing threshold $V_{th}$. Therefore, when the membrane $\mathbf{u}^t_l$ is higher than $V_{th}$, the LIF function fires a spike and the membrane potential is reset to 0. 

\begin{figure*}
    \centering
    \includegraphics[width=18cm]{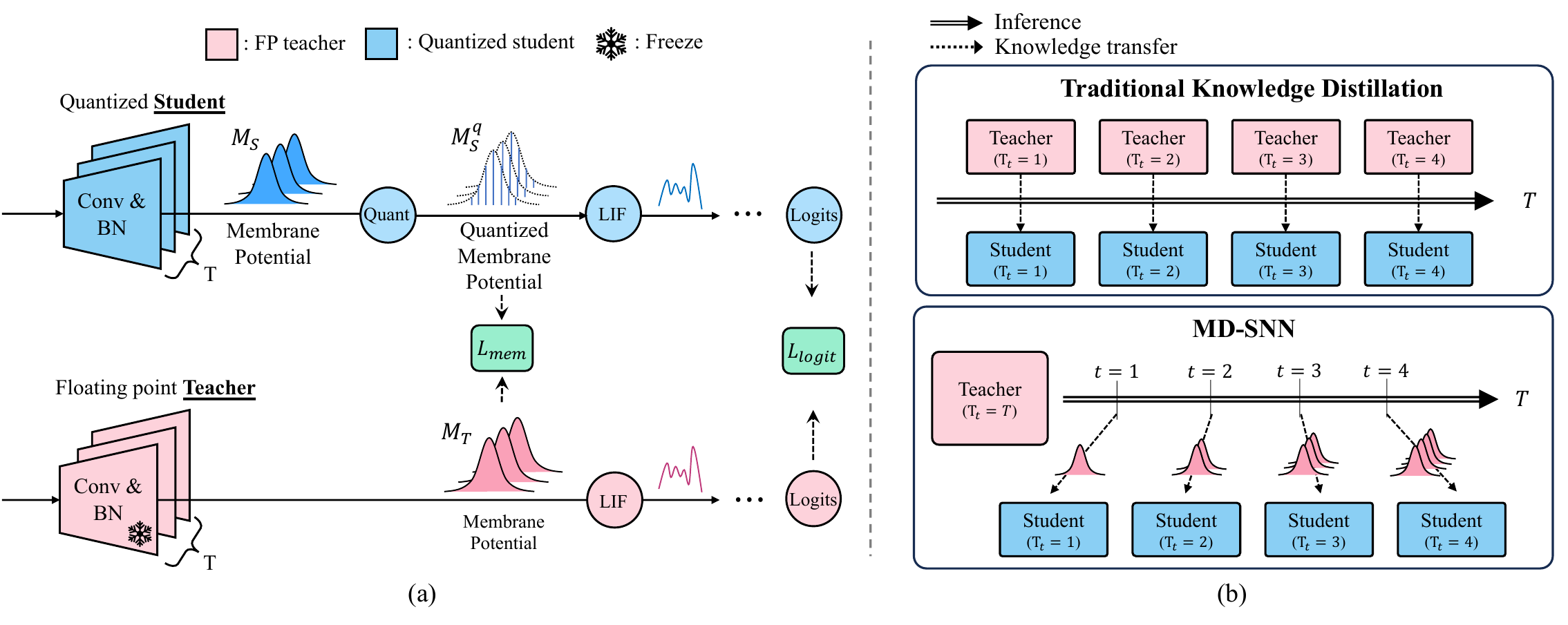}
    \vspace{-3mm}
    \caption{MD-SNN architecture and training paradigm. (a) Dual distillation paths using membrane potentials and logits between FP teacher and quantized student. (b) Versatile teacher framework: single $T=4$ teacher guides multiple students via temporal membrane alignment, contrasted with traditional one-to-one knowledge distillation requiring separate teachers per timestep.}
    \label{fig:main}
    \vspace{-5mm}
\end{figure*}

\subsection{Knowledge Distillation}
Knowledge distillation is a potent technique designed to bridge the gap between larger, well-built teacher models and more compact student models, enabling the students to quickly attain optimal solutions. There are two primary methods in KD: logit distillation and feature distillation. The logit distillation method employs softened logits along with KL divergence, effectively minimizing the distance between the logits of the teacher and student models \cite{hinton2015distilling}. This approach encourages the student model to inherit the knowledge captured within the teacher model's predictions. On the other hand, feature distillation takes a different route. It aims to align the hidden features of both teacher and student models, facilitating rapid convergence and enhancing the student model's performance \cite{heo2019comprehensive}. Our MD-SNN is a branch of feature distillation because we align the membrane potential of the student and the teacher.  
Critically, membrane potentials ($\mathbf{u}^t_l$ in Eqn.\eqref{lif1}) represent pre-activation states before the LIF function, capturing rich temporal dynamics that directly influence spike generation.
Furthermore, we adopt the KL divergence function for membrane distillation instead of the commonly used MSE employed when measuring the distance between hidden features in the teacher and student architectures.

\vspace{-3mm}
\section{Proposed Method}

\subsection{Quantization Schemes}

MD-SNN framework employs distinct quantization strategies for different network components to maintain computational efficiency while preserving information flow. We apply uniform quantization to weights and membrane potentials, and adopt a different approach for Batch Normalization (BN) parameters.

\subsubsection{Weight and Membrane Potential Quantization}

Inspired by MINT \cite{yinmint}, we employ symmetric uniform quantization with adaptive scaling for both weights and membrane potentials. Given a tensor $w$ and target bit-width $b$, our quantization applies a hyperbolic tangent transformation followed by uniform quantization:
\begin{equation}
Q(w, b) = \frac{\alpha \cdot \text{Round}(\bar{w} \cdot \Delta(b))}{\Delta(b)},
\label{quan}
\end{equation}
where $\bar{w} = \text{clamp}(\tanh(w)/\alpha, -1, 1)$, $\alpha = \max(|\tanh(w)|)$ is the dynamic scale factor, and $\Delta(b) = 2^{b-1} - 1$ represents the quantization levels. Here, $\text{clamp}(x, a, b) = \min(\max(x, a), b)$ constrains values to the range $[a, b]$, and $\max(|\cdot|)$ denotes the maximum absolute value across all tensor elements. We employ the Straight-Through Estimator (STE) for gradient backpropagation during training. The $\tanh$ transformation bounds weight values while maintaining smooth gradients, and the adaptive scaling factor $\alpha$ ensures efficient utilization of quantization levels. This approach is particularly effective for membrane potentials, as the adaptive scaling and symmetric quantization provide effective representation of the temporal dynamics in spiking neurons.








\subsubsection{Batch Normalization Quantization}

Batch normalization layers require special consideration as they contain multiple learnable parameters (scale $\gamma$, shift $\beta$) and running statistics (mean $\mu$, variance $\sigma^2$). Direct quantization of these parameters can lead to significant accuracy degradation due to their different value ranges and sensitivities. Following the WAGEUBN approach \cite{yang2020training}, we quantize the weights and bias of BN layers as follows:
\begin{equation}
Q_{BN}(w, b) = \text{clamp}[Q(w, b), -1 + \frac{1}{\Delta(b)+1}, 1 - \frac{1}{\Delta(b)+1}],
\end{equation}
\noindent where $Q(w, b)$ is the quantization function defined in Eq.~\eqref{quan} with $\alpha=1$. The clipping range is slightly reduced from $[-1, 1]$ to ensure proper gradient flow during backpropagation. 


\begin{table}[t]
\centering
\caption{Accuracy comparison of MD-SNN against floating-point baselines on CIFAR-100, N-Caltech101, and TinyImageNet datasets. FP32 refers to full-precision (32-bit) trained without quantization. MS-N denotes our MD-SNN with N-bit quantization for both weights and membrane potentials.}
\begin{NiceTabular}{c|c|c|c||c}
\hline
Dataset & Model & Timestep & Method & Accuracy \\
\hline
\multirow{3}{*}{CIFAR-100} & \multirow{3}{*}{ResNet18} & \multirow{3}{*}{4} 
& FP32 & 74.34\% \\
& & & \textbf{MS-8} & \textbf{74.91\%} \\
& & & \textbf{MS-4} & \textbf{74.41\%} \\
\hline
\multirow{3}{*}{N-Caltech101} & \multirow{3}{*}{ResNet18} & \multirow{3}{*}{10} 
& FP32 & 80.31\% \\
& & & \textbf{MS-8} & \textbf{81.73\%} \\
& & & \textbf{MS-4} & \textbf{81.10\%} \\
\hline
\multirow{3}{*}{TinyImageNet} & \multirow{3}{*}{ResNet34} & \multirow{3}{*}{4} 
& FP32 & 60.34\% \\
& & & \textbf{MS-8} & \textbf{60.59\%} \\
& & & \textbf{MS-4} & \textbf{59.90\%} \\
\hline
\end{NiceTabular}
\label{tab:accuracy}
\vspace{-2.mm}
\end{table}

\begin{table}[t]
\caption{Comparisons between existing quantization methods on SNNs. W and U represent the bit-widths of weight and membrane potential. $\dagger$ represents our implementation.} 
\resizebox{0.485\textwidth}{!}{
\begin{NiceTabular}{c|c|c|c|c|c}
\hline
Data& Method& Model& \begin{tabular}[c]{@{}c@{}}Precision\\ (W / U)\end{tabular} & Timestep & Accuracy \\ \hline
\multirow{9}{*}{CIFAR10} & STBP-Quant~\cite{tan2023low}&AlexNet& 8 / 14& 8& 86.65 \% \\ \cline{2-6} 
& STBP-Quant~\cite{tan2023low}&AlexNet& 4 / 10& 8& 84.09 \% \\ \cline{2-6} 
& ST-Quant~\cite{chowdhury2021spatio}&VGG9& 5 / 32& 25& 88.60 \% \\ \cline{2-6} 
& ADMM-Quant~\cite{deng2021comprehensive}&7Conv+3FC& 4 / 32& 8& 89.40 \% \\ \cline{2-6} 
& SpikeSim~\cite{moitra2023spikesim}&VGG9& 8 / 32& 5& 88.11 \% \\ \cline{2-6} 
& MINT~\cite{yinmint}& ResNet19&8 / 8& 4& 91.36 \% \\ \cline{2-6} 
& MINT~\cite{yinmint}& ResNet19&4 / 4& 4& 91.45 \% \\ \cline{2-6} 
& \textbf{MD-SNN (ours)} &ResNet18& \textbf{8 / 8}& \textbf{4}& \textbf{94.27 \%} \\ \cline{2-6} 
& \textbf{MD-SNN (ours)} &ResNet18 & \textbf{4 / 4}& \textbf{4}& \textbf{94.05 \%} \\ \hline \hline
\multirow{6}{*}{CIFAR100} & ST-Quant~\cite{chowdhury2021spatio}&VGG11& 5 / 32& 30& 66.20 \% \\ \cline{2-6} 
& SpikeSim~\cite{moitra2023spikesim}& VGG16&8 / 32& 10& 65.23 \% \\ \cline{2-6} 
& MINT~\cite{yinmint}$\dagger$&ResNet19& 8 / 8& 4& 68.53 \% \\ \cline{2-6} 
& MINT~\cite{yinmint}$\dagger$&ResNet19& 4 / 4& 4& 68.79 \% \\ \cline{2-6} 
& \textbf{MD-SNN (ours)} &ResNet18& \textbf{8 / 8}& \textbf{4}& \textbf{74.91 \%} \\ \cline{2-6} 
& \textbf{MD-SNN (ours)} &ResNet18& \textbf{4 / 4}& \textbf{4}& \textbf{74.41 \%} \\ \hline \hline
\multirow{5}{*}{TinyImageNet} & SpikeSim~\cite{moitra2023spikesim}& VGG16&8 / 32& 10& 54.03 \% \\ \cline{2-6} 
& MINT~\cite{yinmint}$\dagger$&ResNet34& 8 / 8& 4& 42.94 \% \\ \cline{2-6} 
& MINT~\cite{yinmint}$\dagger$&ResNet34& 4 / 4& 4& 44.94 \% \\ \cline{2-6} 
& \textbf{MD-SNN (ours)} &ResNet34& \textbf{8 / 8}& \textbf{4}& \textbf{60.59 \%} \\ \cline{2-6} 
& \textbf{MD-SNN (ours)} &ResNet34& \textbf{4 / 4}& \textbf{4}& \textbf{59.90 \%} \\ \hline

\end{NiceTabular}
}

\label{comparison}
\vspace{-6mm}
\end{table}

\begin{figure}[t]
    \centering
    \includegraphics[width=85mm]{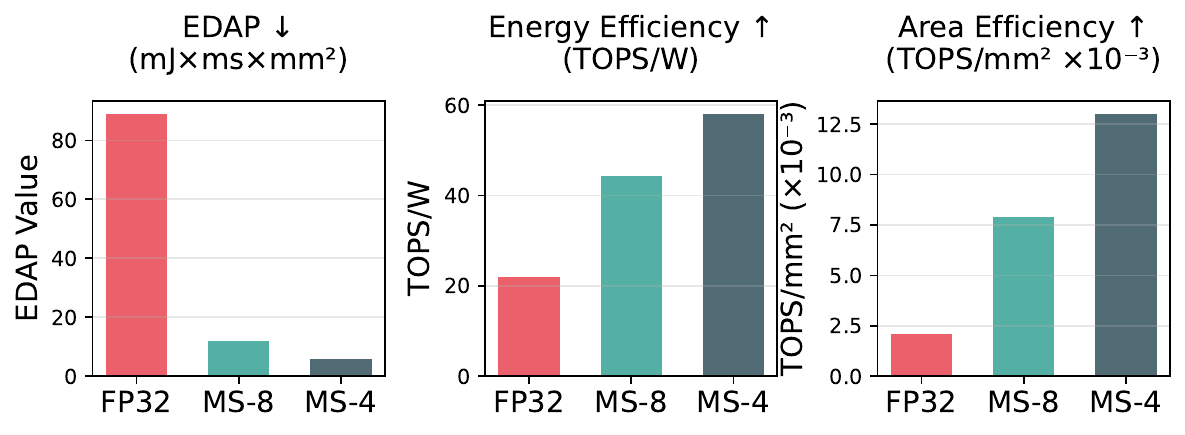}
    \caption{Hardware efficiency comparison of floating-point and MD-SNN approaches on the N-Caltech101 dataset. Lower EDAP values indicate better overall hardware efficiency, while higher TOPS/W and TOPS/mm² values indicate better energy and area efficiency, respectively. FP32 refers to full-precision, MS-N denotes MD-SNN with N-bit quantization for both weights and membrane potentials.}
    \label{fig:hardware}
    \vspace{-6mm}
\end{figure}

\vspace{-1.5mm}
\subsection{Membrane-aware Distillation}
In this section, we present the details of membrane-aware distillation and training strategies. The membrane potential, computed by weights and input spikes shown as Eqn. \eqref{lif1}, \eqref{lif2}, is an important intrinsic component as it determines the output spikes of the specific neuron. Therefore, the membrane distribution of the teacher model can be a good guide to help the student achieve optimal solutions. The holistic scheme of MD-SNN is shown in Fig. \ref{fig:main}(a). We group-wisely and time-wisely extract the membrane distributions and compute the distance of membrane distribution between the teacher and student model as follows:
\vspace{-3.5mm}
\begin{gather}
    L_{mem} = \sum^{group}_{i=0} \sum^{time}_{j=0} D(M_{T, (i,j)}, M^q_{S, (i,j)}) \\
    D(a, b) = \tau_a\tau_b \times KL(\text{log-softmax}(a/\tau_a), \text{softmax}(b/\tau_b)),   
\end{gather}
\noindent where $group$, $time$ are the total number of groups in a ResNet model, timesteps respectively, $M_T$, $M^q_S$ are extracted membrane potential from teacher and quantized student model, $KL$ is KL divergence, and $\tau_a$, $\tau_b$ are temperature parameters of $a$, $b$ in KL divergence. The decision for group-wise distillation is discussed in Section \ref{group-wise}.  The grand goal of $L_{mem}$ is to minimize the distribution similarities between the teacher and student model. The final loss $L$ for training the student model is shown below.
\vspace{-2mm}
\begin{gather}
    L = \alpha L_{CE} + \beta L_{logit} + \gamma L_{mem}\\
    L_{CE} = \text{Cross-Entropy}(z^q_s) \\
    L_{logit} = D(z_t, z^q_s).
\label{loss}
\end{gather}
\noindent Here, $L_{CE}$, $L_{logit}$ represent the Cross-Entropy loss of the student model and logit distillation loss respectively, $z_t, z^q_s$ are the logits of the teacher model and quantized student model respectively. The $\alpha$, $\beta$, and $\gamma$ are the hyperparameters of each component of the loss function.

\subsection{Teacher Versatility}

A distinctive advantage of MD-SNN is exploiting SNN's temporal computation for efficient knowledge transfer. Since a teacher model trained with $T$ timesteps inherently computes membrane potentials at all intermediate timesteps $t \in{1, 2, ..., T}$, a single teacher can guide multiple students with different temporal configurations. As shown in Fig.~\ref{fig:main}(b), we extract membrane potentials at timestep $t$ from a $T=4$ teacher to supervise students configured for $t \leq T$ timesteps. This one-to-many paradigm eliminates training separate teachers for each configuration. The efficiency gain is substantial: instead of 2N model trainings (N teachers + N students), our approach requires only N+1 trainings (1 teacher + N students), significantly reducing training costs.


\section{Experiments}
\subsection{Experimental Setup}
\noindent\textbf{Software Evaluation Setup:} We verify our proposed method on both static and dynamic datasets, e.g., CIFAR10/100, N-Caltech101, and TinyImageNet with ResNet architectures. Our baseline of SNN-based ResNet architecture is MS-ResNet \cite{hu2021advancing} and we use direct coding to convert a float pixel value into binary spikes~\cite{wu2019direct}. While the teacher model is trained with FP values, the weights and membrane potentials of students are quantized to 4 or 8-bit. We also use 8-bit BN parameters in the student model. We set $\alpha, \beta, \gamma$ in Eqn. \eqref{loss} as 1, 3, 1 respectively. We set the number of epochs as 200 for all datasets. The CIFAR10/100, N-Caltech experiments are conducted on RTX2080ti GPUs and the TinyImageNet results are obtained from A100 GPU. For inference evaluation, we use the following hardware evaluation method. 

\noindent \textbf{Hardware Evaluation Setup}: All the baselines and MD-SNN, optimized SNNs are evaluated on SpikeSim, an end-to-end inference evaluation platform based on in-memory computing architectures for SNNs. We use the Spikesim code available in \cite{moitra2023spikesim} to benchmark the baseline as well as the MD-SNN model's energy, delay, area and performance metrics. Tab.~\ref{tab:hw_params} shows the detailed hardware parameters used to initialize the SpikeSim framework. 

Note that the results shown in the tables and figures in the below sections for MD-SNN correspond to the final student model's accuracy or energy obtained with membrane potential distillation.
\begin{table}[h!]
    \caption{Hardware Implementation Parameters.}
    \label{tab:hw_params}
    \centering
    \begin{NiceTabular}{|c|c|} \hline
       Technology & 32nm CMOS \\ \hline
       Crossbar Size  & 64 \\ \hline
       Device & 4-bit FeFET ($\sigma$/$\mu$=20\%) \\ \hline
       $R_{off}$/$R_{on}$ & 10 at $R_{on}$=20k$\Omega$ \\ \hline
       $V_{DD}$ \& $V_{read}$ & 0.9V \& 0.1V \\ \hline
    \end{NiceTabular}
    \vspace{-3mm}
\end{table}


\begin{figure}[t]
    \centering
    \includegraphics[width=90mm]{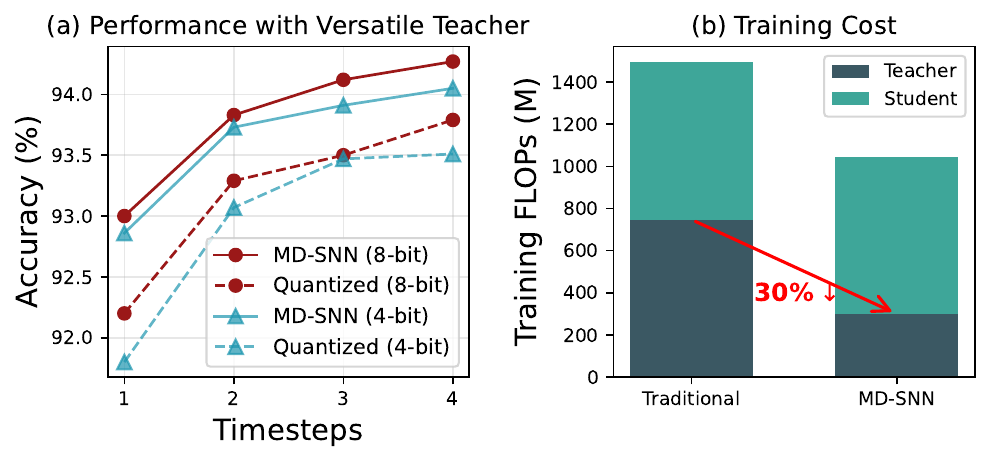}
    \caption{Teacher versatility on CIFAR-10. (a) Accuracy improvements of MD-SNN over quantized baselines using a single $T=4$ teacher for all timestep configurations. Quantized baselines (dashed line) are models trained with MINT~\cite{yinmint} for $t={1,2,3,4}$ timesteps individually without any distillation. MD-SNN results (solid line) undergo distillation. (b) Training FLOPs comparison showing 30\% reduction with the versatile teacher approach. Traditional refers to individual timestep teacher-student distillation (see Fig. 2(b)).}
    \label{fig:versal}
    \vspace{-3mm}
\end{figure}

\subsection{Accuracy and Energy-efficiency Results}
\noindent\textbf{Software Results: }Overall performance results of CIFAR100, N-Caltech101, and TinyImageNet are shown in Tab.~\ref{tab:accuracy}. We use ResNet-18 for CIFAR100 and N-Caltech101, and ResNet-34 for TinyImageNet. The conventional FP models are our comparison baseline. On CIFAR100 and N-Caltech101 datasets, our proposed distilled model's accuracies with 4 and 8 bits are higher than the results of the FP model. In the large dataset, e.g., TinyImageNet, our distilled accuracy achieves competitive performance with the accuracy of the FP architecture.


\noindent\textbf{Hardware Results:} Fig.~\ref{fig:hardware} provides the SpikeSim evaluation, demonstrating significant efficiency gains on the N-Caltech101 dataset. MD-SNN with 4-bit quantization achieves 14.85$\times$ lower energy-delay-area product (EDAP), 2.64$\times$ higher TOPS/W and 6.19$\times$ higher TOPS/mm$^2$ compared to 32-bit baselines at iso-accuracy. The 8-bit variants show intermediate improvements. Notably, our 4-bit models significantly surpass FP models' efficiency while maintaining higher accuracy, validating that membrane distillation effectively compensates for aggressive quantization.

\subsection{Comparison to Previous Works}
To validate the performance of MD-SNN, we compare our work to previous quantization methods on SNNs, including STBP-Quant~\cite{tan2023low}, ST-Quant~\cite{chowdhury2021spatio}, ADMM-Quant~\cite{deng2021comprehensive}, SpikeSim~\cite{moitra2023spikesim}, and MINT~\cite{yinmint} shown in Tab.~\ref{comparison}. We compare the results of quantized SNN architecture on CIFAR10, CIFAR100, and TinyImageNet datasets with different bit-width of weight and membrane potential. Across all datasets, our MD-SNN achieves higher or comparable accuracy than other prior quantized SNN works.

\begin{figure}[t]
    \centering
    \includegraphics[width=70mm]{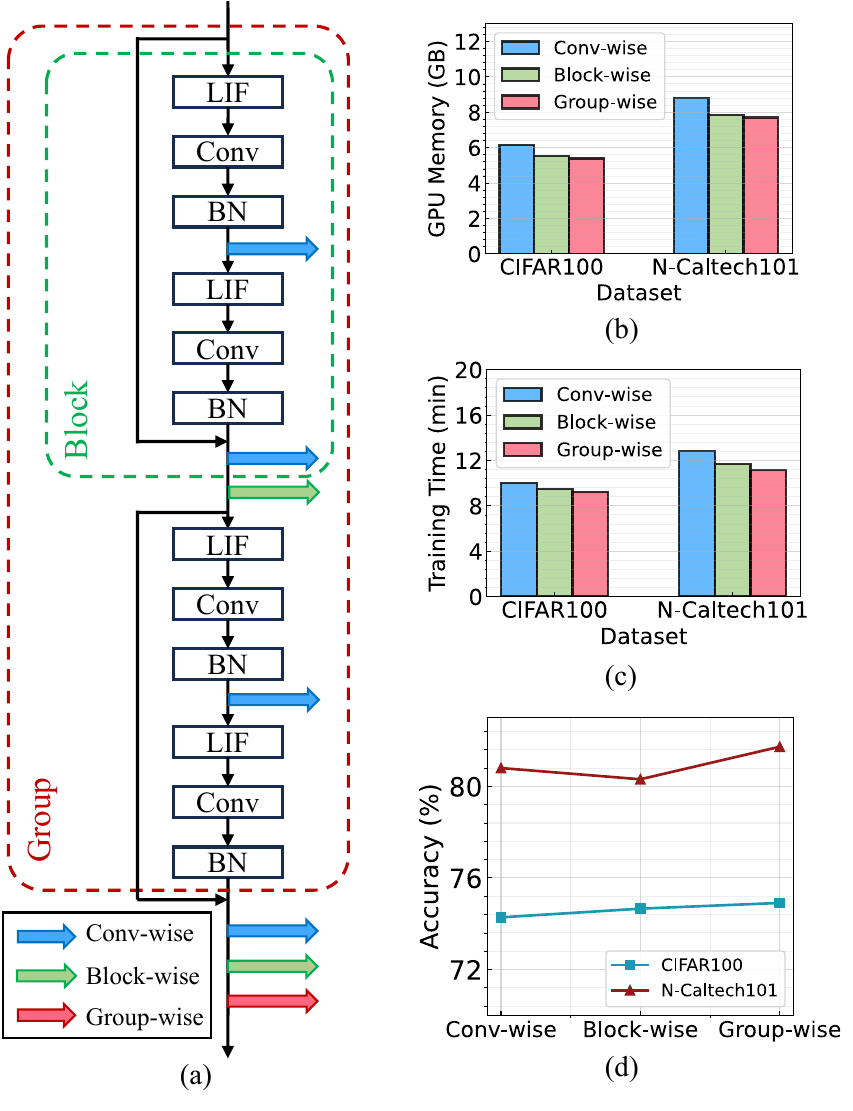}
    \caption{Ablation study on membrane distillation granularity. (a) Three extraction strategies for membrane potential: Conv-wise (after each Conv-BN layer), Block-wise (after each residual block), and Group-wise (after each stage). (b) GPU memory consumption and (c) training time comparison across datasets. (d) Accuracy comparison on CIFAR-100 and N-Caltech101.}
    \label{fig:group_wise}
    \vspace{-3mm}
\end{figure}

\subsection{Effectiveness of Versatile Teacher}

Fig.~\ref{fig:versal} demonstrates the versatile teacher framework on CIFAR-10. A single $T=4$ teacher successfully guides students at all timesteps ($t=1,2,3,4$), achieving 0.48-1.06\% accuracy improvement over quantized models while reducing the total training FLOPs by 30\%. The efficiency gain comes from reusing intermediate membrane potentials. The teacher transfers the knowledge for all $t\leq4$ during forward propagation, eliminating the need for separate teachers per timestep configuration.


\begin{figure}[t]
    \centering
    \subfloat[]{\includegraphics[width=2.8cm]{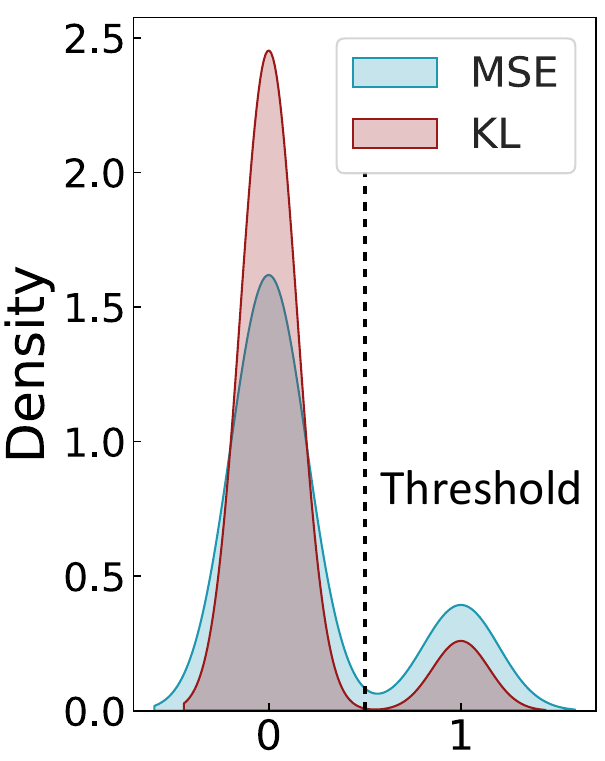}}
    \hspace{1mm}
    \subfloat[]{\includegraphics[width=4.3cm]{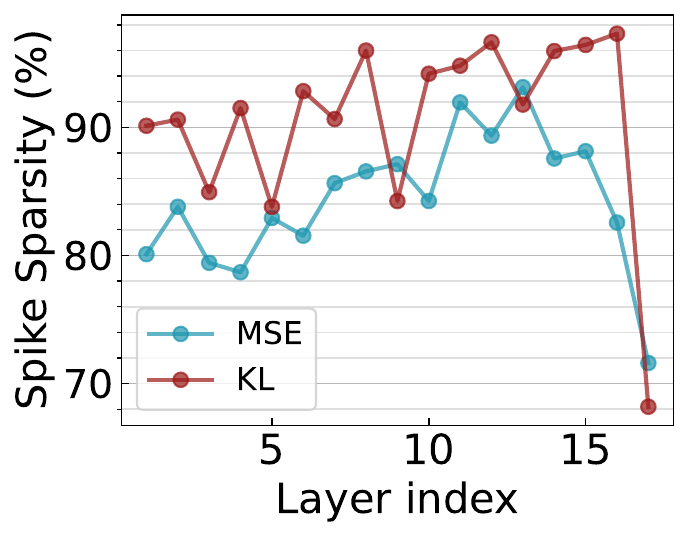}}

    \caption{Spike Sparsity and distribution of membrane potential of a student model trained on N-Caltech101 dataset on ResNet18 architecture. (a) Membrane potential distribution of a student model trained with KL-divergence and MSE distillation loss. The dashed line represents the threshold $V_{th}$. (b) The sparsity according to the layer index.}
    \label{Fig_KLMSE}
\end{figure}

\begin{table}[]
\centering
\caption{The accuracy and sparsity comparison for varying loss methods on membrane distillation. X/Y represent the results when the weights and membrane potential are quantized to 8-bit/4-bit respectively.}
\begin{NiceTabular}{c|c|c|c}
\hline
Dataset& Loss & Accuracy (\%) & Sparsity (\%) \\ 
\hline
\multirow{2}{*}{CIFAR100}& CE+Logit+KL   & 74.91 / 74.41& 83.08 / 81.51\\
& CE+Logit+MSE  & 73.83 / 73.09& 80.00 / 77.78\\ 
\hline
\multirow{2}{*}{N-Caltech101} & CE+Logit+KL   & 81.73 / 81.18& 89.64 / 90.33\\
& CE+Logit+MSE  & 80.74 / 78.01& 83.09 / 82.61\\
\hline
\end{NiceTabular}
\label{Tab_KLMSE}
\vspace{-4mm}
\end{table}

\section{Ablation Study}
\subsection{Group-wise Membrane Distillation}
\label{group-wise}
Given the presence of membrane potential before every LIF neuron, selecting the extraction point for membrane potential distillation is crucial. We propose three distinct approaches: Conv-wise, Block-wise, and Group-wise, as depicted in Fig.~\ref{fig:group_wise}(a). In the Conv-wise approach, we extract the membrane potential after every Convolutional (Conv) and BN layer, transferring this knowledge to the membrane potential of the student model. Alternatively, the block-wise approach involves using the membrane potential immediately after each block, which comprises two Conv layers, two BN layers, and LIF neurons. Lastly, we introduce the group-wise approach, where a group consists of multiple blocks. For example, in ResNet34, a group may consist of 3, 4, or 6 blocks, and the membrane potential knowledge is transferred after each group.

To select the optimal membrane potential distillation method, we analyze GPU memory, training time, and accuracy on CIFAR-100 and N-Caltech101 datasets, as depicted in Fig.~\ref{fig:group_wise}(b)-(d). Group-wise distillation proves most efficient in both GPU memory and training time. Moreover, it achieves the highest accuracy compared to Conv-wise and Block-wise methods. This highlights the importance of balanced knowledge transfer from the teacher model, as excessive teaching can degrade student model performance\cite{mirzadeh2020improved}.



\subsection{KL Divergence vs. MSE for Membrane Distillation}

We investigate the choice of loss function for membrane potential distillation between teacher and student models. Unlike conventional ANN feature distillation that employs MSE~\cite{heo2019comprehensive,romero2014fitnets}, we adopt KL divergence for membrane distribution alignment. Table~\ref{Tab_KLMSE} compares the effectiveness of both approaches on CIFAR-100 and N-Caltech101 datasets, demonstrating superior accuracy when using KL divergence over MSE.

Beyond accuracy improvements, KL divergence also enhances spike efficiency. We measure spike sparsity as the ratio of total spikes across $T$ timesteps to the total number of neurons in the network. Models trained with KL divergence consistently achieve higher sparsity than those using MSE. This difference is evident in the membrane potential distributions shown in Fig.~\ref{Fig_KLMSE}(a). MSE produces a larger density area above the firing threshold compared to KL divergence, indicating excessive spike generation. Consequently, as illustrated in Fig.~\ref{Fig_KLMSE}(b), KL divergence maintains higher spike sparsity across most network layers. This distinction is crucial: MSE loss generates redundant spikes that degrade both computational efficiency and model performance, while KL divergence preserves the sparse, event-driven nature essential for efficient SNN computation.

\vspace{-1mm}
\subsection{Distillation Components}

To validate the importance of membrane distillation for teacher versatility, we compare different distillation strategies in Tab.~\ref{tab:distill_comp}. Using a single $T=4$ teacher to guide students at various timesteps on CIFAR-10 with 4-bit quantized weights and membrane potentials, we evaluate three configurations: quantized model without distillation, logit-only distillation, and our complete MD-SNN (logit+membrane distillation). Full usage of logit and membrane distillation achieves higher improvements across all timesteps compared to logit-only distillation. This demonstrates that membrane potentials, not just logits, enable effective knowledge transfer from a versatile teacher to temporally-diverse students.

\begin{table}[t]
\centering
\caption{Impact of distillation components on teacher versatility.}
\label{tab:distill_comp}
\begin{tabular}{l|cccc}
\hline
Method & T=1 & T=2 & T=3 & T=4 \\
\hline
Quantized (no distillation) & 91.80 & 93.07 & 93.47 & 93.51 \\
Logit-only & 92.15 & 93.21 & 93.55 & 93.78 \\
Logit+Membrane (Ours) & \textbf{92.86} & \textbf{93.73} & \textbf{93.91} & \textbf{94.05} \\
\hline
\end{tabular}
\vspace{-4mm}
\end{table}

\section{Conclusion}
We present MD-SNN, the first knowledge distillation framework that leverages membrane potential dynamics to guide quantized SNN training. By transferring membrane distributions from full-precision teachers to quantized students using KL divergence, our method addresses the critical accuracy degradation problem in SNN quantization. MD-SNN achieves competitive performance over FP baselines across CIFAR-10/100, N-Caltech101, and TinyImageNet datasets, while delivering better efficiency on hardware. Furthermore, the versatile teacher framework, where a single teacher trained at $T=4$ timesteps can effectively guide students at any $t \leq 4$, reduces training costs by 30\% while maintaining superior performance at every operating point. This enables flexible deployment across different latency-accuracy requirements without retraining multiple models.

\vspace{-1mm}
\section*{Acknowledgment}
This work was supported in part by CoCoSys, a JUMP2.0 center sponsored by DARPA and SRC, the National Science Foundation (CAREER Award, Grant \#2312366, Grant \#2318152), the DARPA Young Faculty Award, the DoE MMICC center SEA-CROGS (Award \#DE-SC0023198) and the Global Industrial Technology Cooperation Center (GITCC) program.

\clearpage
\bibliography{references}
\bibliographystyle{unsrt}

\end{document}